\documentclass[10pt,twocolumn,letterpaper]{article}

\usepackage{cvpr}
\usepackage{times}
\usepackage{epsfig}
\usepackage{graphicx}
\usepackage{amsmath}
\usepackage{amssymb}
\usepackage{mathrsfs} % For LaTeX2e
\usepackage{amsmath,bm}
\usepackage{mdwlist}
\usepackage{array,multirow}
\usepackage{paralist}
\usepackage{bbding}
\usepackage{enumerate}
\usepackage{booktabs}
\usepackage{url}
\usepackage{caption}
\usepackage{xspace}
\usepackage{supertabular}
\usepackage{threeparttable}

\usepackage[breaklinks=true,bookmarks=false]{hyperref}

\cvprfinalcopy % *** Uncomment this line for the final submission

\def\cvprPaperID{****} % *** Enter the CVPR Paper ID here

\begin{document}

%%%%%%%%% TITLE
\title{Drone-based Joint Density Map Estimation, Localization and Tracking with Space-Time Multi-Scale Attention Network}

\author{Longyin Wen$^{1}$\thanks{Both authors contributed equally to this work.}, Dawei Du$^{2\ast}$, Pengfei Zhu$^3$, \\
Qinghua Hu$^3$, Qilong Wang$^3$, Liefeng Bo$^1$, Siwei Lyu$^2$\\
$^1$JD Finance America Corporation, Mountain View, CA, USA.\\
$^2$University at Albany, State University of New York, Albany, NY, USA. \\
$^3$Tianjin University, Tianjin, China. \\
{\tt\small longyin.wen.cv@gmail.com, cvdaviddo@gmail.com,} \\
{\tt\small \{zhupengfei, huqinghua, qlwang\}@tju.edu.cn, liefeng.bo@jd.com}
}

\maketitle
%\thispagestyle{empty}

%%%%%%%%% ABSTRACT
\begin{abstract}
This paper proposes a space-time multi-scale attention network (STANet) to solve density map estimation, localization and tracking in dense crowds of video clips captured by drones with arbitrary crowd density, perspective, and flight altitude. Our STANet method aggregates multi-scale feature maps in sequential frames to exploit the temporal coherency, and then predict the density maps, localize the targets, and associate them in crowds simultaneously. A coarse-to-fine process is designed to gradually apply the attention module on the aggregated multi-scale feature maps to enforce the network to exploit the discriminative space-time features for better performance. The whole network is trained in an end-to-end manner with the multi-task loss, formed by three terms, \ie, the density map loss, localization loss and association loss. The non-maximal suppression followed by the min-cost flow framework is used to generate the trajectories of targets' in scenarios. Since existing crowd counting datasets merely focus on crowd counting in static cameras rather than density map estimation, counting and tracking in crowds on drones, we have collected a new large-scale drone-based dataset, DroneCrowd, formed by $112$ video clips with $33,600$ high resolution frames (\ie, $1920\times1080$) captured in $70$ different scenarios. With intensive amount of effort, our dataset provides $20,800$ people trajectories with $4.8$ million head annotations and several video-level attributes in sequences. Extensive experiments are conducted on two challenging public datasets, \ie, Shanghaitech and UCF-QNRF, and our DroneCrowd, to demonstrate that STANet achieves favorable performance against the state-of-the-arts. The datasets and codes can be found at \url{https://github.com/VisDrone}.
\end{abstract}

%%%%%%%%% BODY TEXT
\section{Introduction}
Drones, or general unmanned aerial vehicles (UAVs), equipped with cameras have been fast deployed to a wide range of applications, such as video surveillance for crowd control and public safety \cite{DBLP:journals/cm/MotlaghBT17}. In recent years, many massive stampedes have taken place around the world that claimed many victims, making the automatic density map estimation, counting and tracking in crowds on drones an important task. These tasks have recently drawn great attention from the computer vision research community.

Despite great progress has been achieved in recent years, these algorithms still have room for improvement to deal with video sequences captured by drones, due to various challenges, such as view point and scale variations, background clutter, and small scales. Developing and evaluating crowd counting and tracking algorithms for drones are impeded by the lack of publicly available large-scale benchmarks. Some recent efforts \cite{DBLP:conf/cvpr/ZhangZCGM16,DBLP:conf/eccv/IdreesTAZARS18,DBLP:conf/wacv/ZhangSC18} have devoted to construct datasets for crowd counting. However, these datasets are still limited in sizes and scenarios covered. They only focus on crowd counting with still images by surveillance cameras, due to difficulties in data collection and annotation for drone-based crowd counting and tracking.

To fill this gap, we collect a large-scale drone-based dataset for density map estimation, crowd localization and tracking. Our DroneCrowd dataset consists of $112$ video clips formed by total $33,600$ frames with a resolution of $1920\times1080$ pixels, captured by various drone-mounted cameras, for $70$ different scenarios across $4$ different cities in China (\ie, Tianjin, Guangzhou, Daqing, and Hong Kong). These video clips are manually annotated with more than $4.8$ million head annotations and several video-level attributes. To the best of our knowledge, this is the largest and most thoroughly annotated density map estimation, localization, and tracking dataset to date, see Table \ref{tab:dataset-comparison}.

\begin{table*}[t]
\caption{Comparison of the DroneCrowd dataset with existing datasets.}
\vspace{-2mm}
\centering
\footnotesize \setlength{\tabcolsep}{6.0pt}
\begin{threeparttable}
\begin{tabular}{c|ccccccccc}
\hline
Dataset  &Type  &Resolution  &Frames &Max count &Min count &Ave count &Total count &Trajectory &Year \\
\hline
UCF\_CC\_50 \cite{DBLP:conf/cvpr/IdreesSSS13}    &image   &-      &$50$      &$4,543$      &$94$          &$1,279.5$  &$63,974$  & &2013 \\
Shanghaitech A \cite{DBLP:conf/cvpr/ZhangZCGM16} &image &-     &$482$    &$3,139$ &$33$    &$501.4$       &$241,677$   & &2016 \\
Shanghaitech B \cite{DBLP:conf/cvpr/ZhangZCGM16} &image &$768\times1024$     &$716$ &$578$  &$9$   &$123.6$  &$88,488$ & &2016 \\
AHU-Crowd \cite{DBLP:journals/jvcir/HuCNWL16}       &image    &$576\times720$   &$107$ &$2,201$   &$58$    &$420.6$      &$45,000$ & &2016 \\
CARPK \cite{DBLP:conf/iccv/HsiehLH17}      &image    &$1280\times720$   &$1,448$ &$188$   &$1$            &$62.0$      &$89,777$ & &2017 \\
Smart-City  \cite{DBLP:conf/wacv/ZhangSC18}      &image    &$1920\times1080$   &$50$ &$14$   &$1$            &$7.4$      &$369$ & &2018 \\
UCF-QNRF \cite{DBLP:conf/eccv/IdreesTAZARS18}    &image   &- &$1,535$   &$12,865$     &$49$      &$815.4$      &$1,251,642$  & &2018 \\
\hline
\hline
UCSD \cite{DBLP:conf/cvpr/ChanLV08}               &video   &$158\times238$   &$2,000$ &$46$          &$11$         &$24.9$       &$49,885$  & &2008 \\
Mall \cite{DBLP:conf/iccv/LoyGX13}       &video    &$640\times480$   &$2,000$ &$53$   &$13$       &$31.2$      &$62,316$ & &2013 \\
WorldExpo  \cite{DBLP:conf/cvpr/ZhangLWY15}\tnote{$\bm{\ast}$}      &video    &$576\times720$   &$3,980$ &$253$   &$1$            &$50.2$      &$199,923$ & &2015 \\
FDST \cite{DBLP:conf/icmcs/FangZCGH19} &video    &$1920\times1080$   &$15,000$ &$57$   &$9$            &$26.7$      &$394,081$ & &2019 \\
\hline
\hline
Ours          &video &$1920\times1080$ &$33,600$ &$455$ &$25$ &$144.8$ &$4,864,280$ &$\checkmark$ &2019 \\
\hline
\end{tabular}
\begin{tablenotes}
\item[$\bm{\ast}$] This dataset includes $1,132$ video sequences captured by $108$ surveillance cameras. Only $3,980$ sampled video frames are annotated with a total of $199,923$ labeled heads of people.
\end{tablenotes}

\end{threeparttable}
\label{tab:dataset-comparison}
\end{table*}

To handle this challenging dataset, we design a new space-time multi-scale attention network (STANet) to solve the density map estimation, localization, and tracking on drones. Specifically, we first aggregate multi-scale feature maps in sequential frames to exploit the temporal coherency, and then generate the enhanced space-time multi-scale features for the prediction of density and localization maps as well as association between consecutive frames. Meanwhile, we gradually apply the attention module on the aggregated feature maps to enforce the network to exploit discriminative space-time features for better performance. The whole network is trained in an end-to-end manner with the multi-task loss, formed by three terms, \ie, the density map loss, localization loss and association loss. After that, we use the non-maximal suppression method to localize the targets, which post-processes the localization results for each video frame by finding the local peaks or maximums based on a threshold. The min-cost flow algorithm \cite{DBLP:conf/cvpr/PirsiavashRF11} is further used to associate the nearest localized head points to generate the trajectories of people's heads in the video. Extensive experiments are carried out on two challenging public datasets (\ie, Shanghaitech \cite{DBLP:conf/cvpr/ZhangZCGM16} and UCF-QNRF \cite{DBLP:conf/eccv/IdreesTAZARS18}) and our DroneCrowd dataset to demonstrate the effectiveness of our method.

{\bf Contributions}. (1) We design a space-time multi-scale attention network to solve the density map estimation, localization, and tracking tasks simultaneously, which gradually apply the attention module on the aggregate multi-scale feature maps to enforce the network to exploit discriminative space-time features for better performance. (2) We present a large-scale drone-based dataset for density map estimation, localization, and tracking in dense crowd, which significantly surpasses existing datasets in terms of data type and volume, annotation quality, and difficulty. (3) Extensive experiments are carried out on three challenging datasets to validate the effectiveness of our STANet method.

\begin{figure*}[t]
\centering
\includegraphics[width=1.0\linewidth]{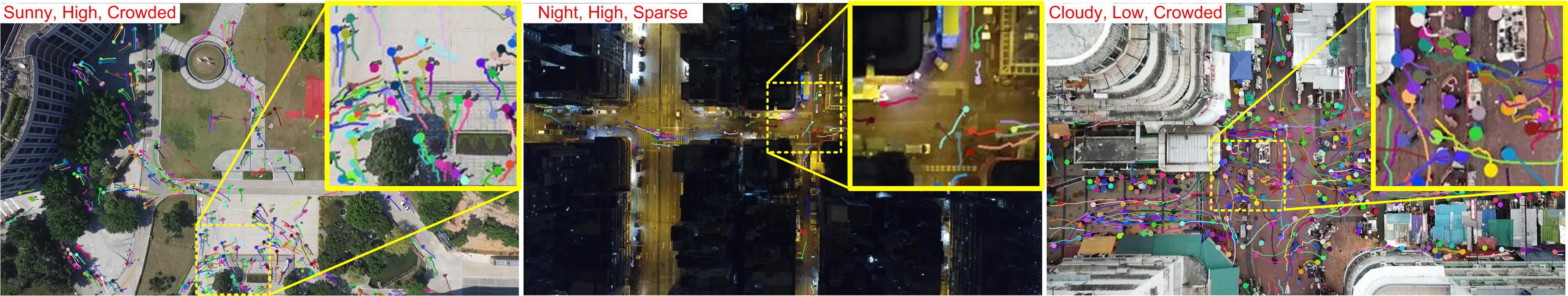}
\caption{Some annotated example frames in the DroneCrowd dataset. Different color indicates different object instance and the corresponding trajectory. The video-level attributes are presented on the top-left corner in each video frame.}
\label{fig:annotations}
\end{figure*}

\section{Related Work}
{\flushleft {\bf Existing datasets.}} To date, there only exists a handful of crowd counting, density map estimation, crowd localization, or crowd tracking datasets. UCF\_CC\_50 \cite{DBLP:conf/cvpr/IdreesSSS13} is formed by $50$ images containing $64,000$ annotated humans, with the head counts ranging from $94$ to $4,543$. Shanghaitech \cite{DBLP:conf/cvpr/ZhangZCGM16} includes $1,198$ images with a total number of $330,165$ labeled people. Recently, UCF-QNRF \cite{DBLP:conf/eccv/IdreesTAZARS18} is released with $1,535$ images and $1.25$ million annotated people's heads in various scenarios. Hsieh~\etal \cite{DBLP:conf/iccv/HsiehLH17} present a drone-based car counting dataset, which approximately contains $90,000$ cars captured in different parking lots. However, these datasets are still limited in sizes and scenarios covered, due to the difficulties in data collection and annotation. 

To evaluate counting algorithms in videos, Chan \etal \cite{DBLP:conf/cvpr/ChanLV08} present the UCSD counting dataset with a resolution of $238\times158$. It includes low density crowd and counting difficulty. Mall \cite{DBLP:conf/iccv/LoyGX13} consists of $2,000$ frames with a resolution of $320\times240$. Similar to UCSD, it is collected by the surveillance camera in a single location. Zhang \etal \cite{DBLP:conf/cvpr/ZhangLWY15} present the WorldExpo dataset with $3,980$ annotated video frames in total, which is captured in $108$ different scenes during 2010 Shanghai WorldExpo. Recently, Fang \etal \cite{DBLP:conf/icmcs/FangZCGH19} collect a video dataset with $15,000$ frames and $394,000$ annotated heads captured from $13$ different scenes. In contrast to the aforementioned datasets, our DroneCrowd dataset is a large-scale drone-based dataset for density map estimation, crowd localization and tracking, which consists of $112$ video sequences with more than $4.8$ million head annotations on $20,800$ people trajectories.

{\flushleft {\bf Crowd counting and density map estimation.}}
The majority of early crowd counting methods \cite{DBLP:conf/iccv/WuN05,DBLP:conf/cvpr/LeibeSS05,DBLP:conf/cvpr/WangW11} rely on sliding-window detector to scan still images or video frames to detect the pedestrians based on their hand-crafted appearance features. However, the detector-based methods are easily affected by heavy occlusion, scale and viewpoint variations in crowded scenarios.

Recently, some methods \cite{DBLP:conf/nips/LempitskyZ10,DBLP:conf/cvpr/ZhangZCGM16,DBLP:conf/eccv/Onoro-RubioL16,DBLP:conf/cvpr/SamSB17,DBLP:conf/eccv/CaoWZS18,DBLP:conf/cvpr/LiZC18,DBLP:journals/corr/abs-1811-10452} formulate crowding counting as the estimation of density maps. Lempitsky and Zisserman \cite{DBLP:conf/nips/LempitskyZ10} learn to infer the density estimation by a minimization of a regularized risk quadratic cost function. Zhang \etal \cite{DBLP:conf/cvpr/ZhangZCGM16} use the multi-column CNN network to estimate the crowd density map, which learns the features for different head sizes by each column CNN. O{\~{n}}oro{-}Rubio and L{\'{o}}pez{-}Sastre \cite{DBLP:conf/eccv/Onoro-RubioL16} design the Hydra CNN, which learns a multi-scale non-linear regression model using a pyramid of image patches extracted at multiple scales to generate the final density prediction. Sam \etal \cite{DBLP:conf/cvpr/SamSB17} develop the switching CNN model to handle the variations of crowd density. Cao \etal \cite{DBLP:conf/eccv/CaoWZS18} propose an encoder-decoder network, where the encoder extracts multi-scale features with scale aggregation and the decoder generates high-resolution density maps using transposed convolutions. Li \etal \cite{DBLP:conf/cvpr/LiZC18} employ dilated convolution layers to enlarge receptive fields and extract deeper features without losing resolutions. In contrast to existing methods, our STANet uses a coarse-to-fine process, which sequentially applies the attention module on multi-scale feature maps to enforce the network to exploit discriminative features. 

In terms of crowd counting in videos, spatio-temporal information is critical to improve the counting accuracy. Xiong \etal \cite{DBLP:conf/iccv/XiongSY17} design a convolutional LSTM model to fully capture both spatial and temporal dependencies for crowd counting. Zhang \etal \cite{DBLP:conf/iccv/ZhangWCM17} combine fully convolutional neural networks and LSTM by residual learning to perform vehicle counting. Different from these two methods, our STANet combines multi-scale feature maps in sequential frames and outputs the enhanced features by deformable convolution, which is effective in exploiting the temporal coherency across frames for better performance.

{\flushleft \textbf{Crowd localization and tracking.}} Besides crowd counting, crowd localization and tracking are also important tasks in safety control scenarios. Rodriguez \etal \cite{DBLP:conf/iccv/RodriguezLSA11} formulate an energy minimization framework by jointly optimizing the density and the location of people, with the temporal-spatial constraints of person tracks in video. Ma \etal \cite{DBLP:conf/cvpr/MaYC15} first obtain local counts from sliding windows over the density map and then use integer programming to recover the locations of individual objects. In \cite{DBLP:conf/eccv/IdreesTAZARS18}, crowd counting and localization tasks are simultaneously solved with a CNN model trained by a composition loss.

\section{DroneCrowd Dataset}
\begin{figure*}[t]
\centering
\includegraphics[width=1.0\linewidth]{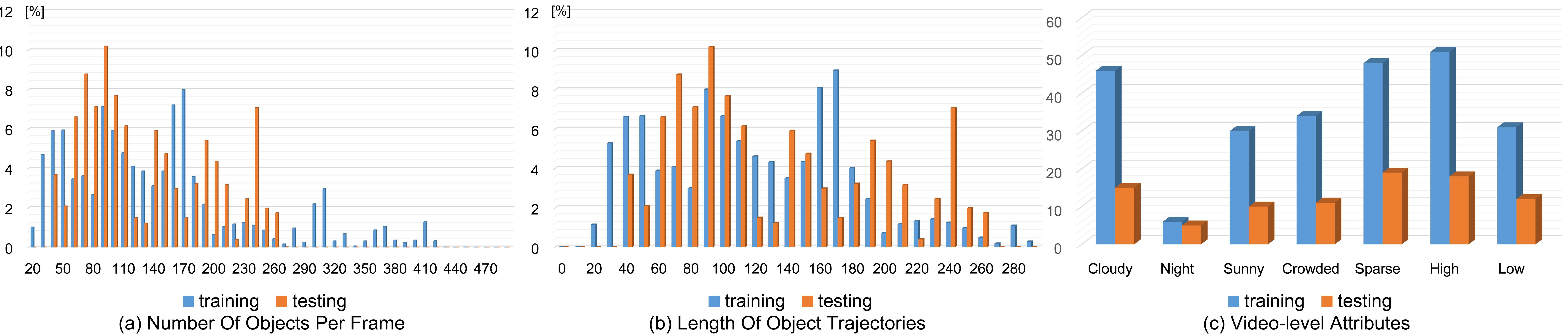}
\caption{(a) The distribution of the number of objects per frame, (b) the distribution of the length of object trajectories, and (c) the attribute statistics, of the {\tt training} and {\tt testing} sets in the DroneCrowd dataset.}
\label{fig:attributes}
\end{figure*}

\subsection{Data Collection and Annotation}
Our DroneCrowd dataset is captured by drone-mounted cameras (DJI Phantom 4, Phantom 4 Pro and Mavic), covering a wide range of scenarios, \eg, campus, street, park, parking lot, playground and plaza. The videos are recorded at $25$ frames per seconds (FPS) with a resolution of $1920\times1080$ pixels. As presented in Figure \ref{fig:attributes} (a) and (b), the maximal and minimal numbers of people in each video frame are $455$ and $25$ respectively, and the average number of objects is $144.8$. More than $20$ thousands of head trajectories of people are annotated with more than $4.8$ million head points in individual frames of $112$ video clips. In terms of annotation, over $20$ domain experts annotate and double-check the dataset using the vatic software \cite{DBLP:journals/ijcv/VondrickPR13} for more than two months. Figure \ref{fig:annotations} shows some frames with annotated trajectories of people heads in video sequences.

We divide the DroneCrowd dataset into the {\tt training} and {\tt testing} sets, with $82$ and $30$ sequences, respectively. Notably, training videos are taken at different locations from testing videos to reduce the chances of algorithms to overfit to particular scenes. The DroneCrowd dataset contains video sequences with large variations in scale, viewpoint, and background clutters. To analyze the performance of algorithms thoroughly, we define three video-level attributes of the dataset, \ie,
\begin{itemize*}
\item \textit{Illumination.} Under different illumination conditions, the objects assume different in appearance. We consider three categories of illumination conditions in DroneCrowd, including {\it Cloudy}, {\it Sunny}, and {\it Night}.
\item \textit{Altitude} is the flying height of drones, which significantly affects the scales of objects. Referring the scales of objects, we define two altitude levels, namely \textit{High} ($<70m$) and \textit{Low} ($>70m$).
\item \textit{Density} indicates the number of objects in each frame. Based on the average number of objects in each frame, we divide the dataset into two density levels, \ie, {\it Crowded} (with the number of objects in each frame larger than $150$), and {\it Sparse} (with the number of objects in each frame less than $150$).
\end{itemize*}
The distribution of video sequences based on the attributes is shown in Figure \ref{fig:attributes} (c).

\subsection{Evaluation Metrics and Protocols}
{\flushleft \textbf{Density map estimation.}}
Following the previous works \cite{DBLP:conf/cvpr/ZhangLWY15,DBLP:conf/cvpr/ZhangZCGM16,DBLP:conf/eccv/IdreesTAZARS18}, the density map estimation task aims to compute per-pixel density at each location in the image, while preserving spatial information about distribution of people. We use the mean absolute error (\text{MAE}) and mean squared error (\text{MSE}) to evaluate the performance, \ie,
\begin{equation}
\begin{aligned}
&\text{MAE} = \frac{1}{\sum_{i=1}^{K}N_i}\sum_{i=1}^{K}\sum_{j=1}^{N_i}{|z_{i,j}-\hat{z}_{i,j}|},\\
&\text{MSE} = \sqrt{\frac{1}{\sum_{i=1}^{K}N_i}\sum_{i=1}^{K}\sum_{j=1}^{N_i}{|z_{i,j}-\hat{z}_{i,j}|^2}},
\end{aligned}
\end{equation}
where $K$ is the number of video clips, $N_i$ is the number of frames in the $i$-th video. $z_{i,j}$ and $\hat{z}_{i,j}$ are the ground-truth and estimated number of people in the $j$-th frame of the $i$-th video clip, respectively. As stated in \cite{DBLP:conf/cvpr/ZhangZCGM16}, \text{MAE} and \text{MSE} describe the accuracy and robustness of the estimation.

{\flushleft \textbf{Crowd localization.}}
According to \cite{DBLP:conf/eccv/IdreesTAZARS18}, the ideal approach for crowd counting is to detect all people in an image and generate the number of detections, which is critical in several applications such as safety and surveillance. Specifically, each evaluated algorithm is required to output a series of detected points with confidence scores for each test image. The estimated localizations determined by the confidence threshold are associated to the ground-truth localizations using greedy method. Then, we compute the mean average precision (L-mAP) at various distance thresholds ($1,2,3,\cdots,25$ pixels) to evaluate the localization results. We also report the performance with three specific distance thresholds, \ie, L-AP$@10$, L-AP$@15$, and L-AP$@20$ pixels. These criteria penalize missing detection of people as well as duplicate detections (\ie, two detection results for the same people).

{\flushleft \textbf{Crowd tracking.}}
Crowd tracking requires an evaluated algorithm to recover the trajectories of people in video sequence. We use the tracking evaluation protocol in \cite{isvrc-2017} to evaluate the algorithms. Specifically, each tracker is required to output a series of head points with confidence scores and the corresponding identities. We sort the tracklets, formed by the detected locations with the same identity, based on the average confidence of their detections. A tracklet is considered to be correct if the matched ratio between the predictions and ground-truth tracklets is larger than a threshold. Similar to \cite{isvrc-2017}, we use three thresholds in evaluation, \ie, $0.10$, $0.15$, and $0.20$. The matching distance threshold between the predicted and ground-truth locations on the tracklets is set to $25$ pixels. The mean average precision (T-mAP) scores over different thresholds (\ie, T-AP$@0.10$, T-AP$@0.15$, and T-AP$@0.20$) are used to measure the performance. Please refer to \cite{isvrc-2017} for more details.

\begin{figure*}[t]
\centering
\includegraphics[width=\linewidth]{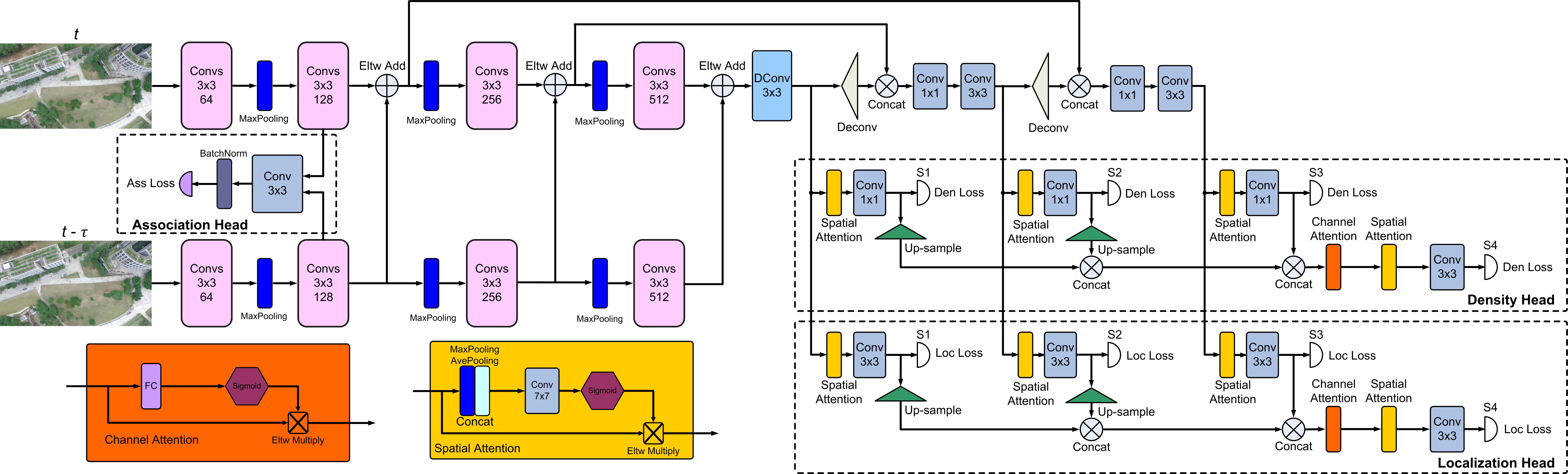}
\caption{The architecture of our space-time multi-scale attention network for crowd counting. The pink rectangles indicate the convolution groups in VGG-16. The light blue rectangle indicates the deformable convolution layer \cite{DBLP:conf/cvpr/ZhuHLD19}.}
\label{fig:architecture}
\end{figure*}

\section{Space-Time Multi-Scale Attention Network}
Our STANet combines multi-scale feature maps in sequential frames, see Figure \ref{fig:architecture}. Meanwhile, we gradually use the attention module on the combined feature maps to enforce the network to focus on the discriminative space-time features. Finally, the non-suppression and min-cost flow association algorithms \cite{DBLP:conf/cvpr/PirsiavashRF11} are used to localize the heads of people and generate their trajectories in video sequence.

{\flushleft \textbf{Network architecture.}}
As shown in Figure \ref{fig:architecture}, our STANet method is constructed on the first four groups of convolution layers in the VGG-16 network \cite{DBLP:journals/corr/SimonyanZ14a}, the backbone network of STANet, to extract the multi-scale features. Motivated by \cite{DBLP:conf/miccai/RonnebergerFB15}, we use the U-Net style architecture to fuse multi-scale features for prediction. Meanwhile, to exploit the temporal coherency, we merge the multi-scale features of the $(t-\tau)$-th frame, and concatenate the features of the $t$-th frame and the $(t-\tau)$-th frame, where $\tau$ is a predefined parameter determining the frame gap between the two frames in temporal coherency\footnote{For the time index $t\leq\tau$, we use the feature of the first frame to exploit the temporal coherency.}. We gradually apply the spatial attention module \cite{DBLP:conf/eccv/WooPLK18} on multi-scale features to enforce the network to focus on the discriminative features (see the black dashed bounding box in Figure \ref{fig:architecture}). After each spatial attention module, one $1\times1$ convolution layer is used to compress the number of channels for efficiency. The multi-scale feature maps of the network are concatenated, merged by the channel and spatial attention modules and one $3\times3$ convolution layer to predict the final density and localization maps. Besides, one $3\times3$ convolution layer is used to exploit the appearance features from the shared backbone in consecutive frames. Then the targets with the same identification are associated based on the normalized features. 

{\flushleft \textbf{Multi-task loss function.}}
The overall loss function consists of density map loss, localization loss and association loss, which is formulated as follows.
\begin{equation}
\begin{array}{ll}
{\cal L} &= \frac{1}{2N}\sum_{n=1}^{N}\big(\lambda_\text{den}{\it L}_\text{den}(\hat{\Phi}^{(n)}, \Phi^{(n)}) \\
&+\lambda_\text{loc}{\it L}_\text{loc}(\hat{\Psi}^{(n)}, \Psi^{(n)})
+\lambda_\text{ass}{\it L}_\text{ass}(\mathcal{D}^{(n)}_\text{s},\mathcal{D}^{(n)}_\text{d})\big),
\end{array}
\label{equ:loss}
\end{equation}
where $N$ is the batch size and $n$ is the index of sample. $\hat{\Phi}^{(n)}$ and $\Phi^{(n)}$ are the estimated and ground-truth density map, while $\hat{\Psi}^{(n)}$ and $\Psi^{(n)}$ are the estimated and ground-truth localization map. $\mathcal{D}^{(n)}_\text{s}$ and $\mathcal{D}^{(n)}_\text{d}$ are the feature distance between the same targets and different targets in consecutive frames, respectively. $\lambda_\text{den}$, $\lambda_\text{loc}$ and $\lambda_\text{ass}$ are balancing factors for the three terms.

Specifically, we use the same pixel-wise Euclidean loss on multi-scale density and localization maps, making different branches (\ie, \text{S1}, \text{S2}, \text{S3}, and \text{S4} in Figure \ref{fig:architecture}) in the network focus on different scales of objects to generate more accurate prediction. For example, the density loss is computed as
\begin{equation}
\begin{array}{ll}
{\cal L}_{\text{den}}=\sum_{s=1}^{S}\sum_{i=1}^{W}\sum_{j=1}^{H}\omega_{s}\cdot\|\Phi(i,j,s) - \hat{\Phi}(i,j,s) \|^2_2,
\end{array}
\label{equ:map_loss}
\end{equation}
where $W$ and $H$ are the width and height of the map, $S$ is the number of scales in the network. $\hat{\Phi}(i,j,s)$ and $\Phi(i,j,s)$ are the estimated and ground-truth density map at pixel location $(i,j)$ of the $n$-th training sample with scale $s$, respectively. $\omega_s$ is the pre-set weight used to balance the losses of different scales of density maps. Notably, following \cite{DBLP:conf/cvpr/ZhangZCGM16}, the geometry-adaptive Gaussian kernel method is used to generate the ground-truth density map $\Phi(i,j,s)$. Similar to \cite{DBLP:journals/corr/abs-1904-07850}, we also generate localization maps using a fixed Gaussian kernel $k$. If the two Gaussians overlap, we take the maximal values.

Inspired by \cite{DBLP:journals/corr/HermansBL17}, we train the association head using the batch hard triplet loss, which samples hard positives and hard negatives for each target. The loss is computed as
\begin{equation}
\begin{array}{ll}
{\cal L}_\text{ass} &= \frac{1}{M}\sum_{i,j=1}^{M}\max\big(\max_{\text{id}_i}\mathcal{D}_\text{s}(i,i) \\ 
&-\min_{id_i\neq id_j}\mathcal{D}_\text{d}(i,j)+\alpha, 0\big),
\end{array}
\label{equ:loss}
\end{equation}
where $\alpha$ is the margin between $\mathcal{D}_\text{s}$ and $\mathcal{D}_\text{d}$ and $M$ is the number of targets in the current frame. Each target with the id $i, j\in M$ contains an association feature.

\begin{table*}[t]
\caption{Comparison of our approach with the state-of-the-art methods on three public datasets.}
\centering
\setlength{\tabcolsep}{15.0pt}
\footnotesize{
\begin{tabular}{c|c|cc|cc|cc}
\hline
\multirow{2}{*}{Method} &\multirow{2}{*}{Venue \& Year} &\multicolumn{2}{c|}{Shanghaitech Part A \cite{DBLP:conf/cvpr/ZhangZCGM16}} &\multicolumn{2}{c|}{Shanghaitech Part B \cite{DBLP:conf/cvpr/ZhangZCGM16}} &\multicolumn{2}{c}{UCF-QNRF \cite{DBLP:conf/eccv/IdreesTAZARS18}} \\
\cline{3-8}
& &$\text{MAE}$ &$\text{MSE}$ &$\text{MAE}$ &$\text{MSE}$ &$\text{MAE}$ &$\text{MSE}$ \\
\hline
MCNN \cite{DBLP:conf/cvpr/ZhangZCGM16} &CVPR 2016        &$110.2$ &$173.2$ &$26.4$ &$41.3$ &$277.0$ &$426.0$ \\
C-MTL \cite{DBLP:conf/avss/SindagiP17} &AVSS 2017               &$101.3$ &$152.4$ &$20.0$ &$31.1$ &$252.0$ &$514.0$ \\
SwitchCNN \cite{DBLP:conf/cvpr/SamSB17} &CVPR 2017          &$90.4$   &$135.0$ &$21.6$ &$33.4$ &$228.0$ &$445.0$ \\
CP-CNN \cite{DBLP:conf/iccv/SindagiP17} &ICCV 2017              &$73.6$   &$106.4$ &$20.1$ &$30.1$ &- &- \\
SaCNN \cite{DBLP:conf/wacv/ZhangSC18} &WACV 2018           &$86.8$   &$139.2$ &$16.2$ &$25.8$ &- &- \\
ACSCP \cite{DBLP:conf/cvpr/ShenXNWHY18} &CVPR 2018      &$75.7$    &$102.7$ &$17.2$ &$27.4$ &- &- \\
IG-CNN \cite{DBLP:conf/cvpr/SamSBS18} &CVPR 2018             &$72.5$    &$118.2$ &$13.6$ &$21.1$ &- &- \\
Deep-NCL \cite{DBLP:conf/cvpr/ShiZLCYCZ18} &CVPR 2018    &$73.5$    &$112.3$ &$18.7$ &$26.0$ &- &- \\
CSRNet \cite{DBLP:conf/cvpr/LiZC18} &CVPR 2018                    &$68.2$   &$115.0$ &$10.6$ &$16.0$ &- &- \\
CL-CNN \cite{DBLP:conf/eccv/IdreesTAZARS18} &ECCV 2018     &- &- &-  &- &$132.0$ &$191.0$  \\
ic-CNN \cite{DBLP:conf/eccv/RanjanLH18} &ECCV 2018         &$68.5$ &$116.2$ &$10.7$  &$16.0$ &- &-  \\
SANet \cite{DBLP:conf/eccv/CaoWZS18} &ECCV 2018            &$67.0$ &$104.5$ &$8.4$ &$13.6$ &- &- \\
SFCN \cite{DBLP:conf/cvpr/WangGL019} &CVPR 2019            &$64.8$ &$107.5$ &$7.6$ &$13.0$ &$102.0$ &$171.4$ \\
ADCrowdNet \cite{DBLP:conf/cvpr/LiuLZNPW19} &CVPR 2019     &${\bf 63.2}$ &${\bf 98.9}$ &$7.6$ &$13.9$ &- &- \\
TEDnet \cite{DBLP:conf/cvpr/JiangXZZ0D019} &CVPR 2019      &$64.2$ &$109.1$ &$8.2$ &$12.8$ &$113.0$ &$188.0$ \\
\hline
\hline
Ours &- &$63.7$ &$101.5$ &${\bf 7.4}$  &${\bf 11.0}$ &$107.6$ &$174.8$  \\
\hline
\end{tabular}}
\label{tab:comparison-results}
\end{table*}

{\flushleft \textbf{Data augmentation.}}
We randomly flip and crop the training images to increase diversity in training data. Due to limited computation resources, for the image size larger than $1920\times1080$, we first resize the image such that its size is smaller than $1920\times1080$. Then we equally divide it into $2\times2$ patches, and use the divided $4$ patches for training.

{\flushleft \textbf{Optimization.}}
In \eqref{equ:loss}, the margin $\alpha$ is set as $0.2$, and the pre-set weights are set as $\lambda_\text{den}=1$, $\lambda_\text{loc}=0.0001$ and $\lambda_\text{ass}=10$ for balance. The pre-set weight in \eqref{equ:map_loss} is set as $\omega=\{0.0125,0.125,0.5,0.5\}$ empirically. The Gaussian normalization method is used to randomly initialize the parameters in the other (de)convolution layers. We set the batch size $N$ to $9$ in training. The network is trained with the learning rate of $10^{-6}$ in the first $10$ epochs, and trained with the learning rate of $10^{-5}$ in the $20$ epochs using the Adam optimization algorithm \cite{DBLP:journals/corr/KingmaB14}.

{\flushleft \textbf{Localization and tracking.}}
After obtaining the localization map of each frame, we use the non-maximal suppression method to localize the heads of people in each frame based on a preset threshold $\theta$. That is, we find the local peaks or maximums density values larger than $\theta$ on the predicted localization map of each video frame to determine the head locations of people. Then, we calculate the Euclidean distance between different pairs of heads in sequential frames and use the min-cost flow algorithm \cite{DBLP:conf/cvpr/PirsiavashRF11} to associate the nearest head points to generate their trajectories. 

\section{Experiments}
We evaluate our method on three challenging datasets. The experiments are conducted on a workstation with Intel E5-2609 CPU, 32GB RAM, and $3$ NVIDIA GeForce GTX 1080Ti GPUs.

\subsection{Public Datasets}
As presented in Table \ref{tab:comparison-results}, we evaluate our STANet method on Shanghaitech \cite{DBLP:conf/cvpr/ZhangZCGM16} and UCF-QNRF \cite{DBLP:conf/eccv/IdreesTAZARS18}. Since they only focus on crowd counting on images, we remove the association head in our STANet method for evaluation.

{\flushleft \textbf{The Shanghaitech dataset}} \cite{DBLP:conf/cvpr/ZhangZCGM16} is formed by $1,198$ images, with a total of $330,165$ annotated people, which is divided into Part A ($482$ images) and Part B ($716$ images). Table \ref{tab:comparison-results} shows the errors of our STANet method as well as $14$ state-of-the-art methods. As show in Table \ref{tab:comparison-results}, our method performs favorably against the state-of-the-arts with $63.7$ $\text{MAE}$ and $101.5$ $\text{MSE}$ in Part A, and $7.4$ $\text{MAE}$ and $11.0$ $\text{MSE}$ in Part B. ADCrowdNet \cite{DBLP:conf/cvpr/LiuLZNPW19} performs better than our method in $\text{MAE}$ ($63.2$ {\em vs.} $63.7$) of Part A, but worse than our method in $\text{MAE}$ ($7.6$ {\em vs.} $7.4$) of Part B.

{\flushleft \textbf{The UCF-QNRF dataset}} \cite{DBLP:conf/eccv/IdreesTAZARS18} contains $1,535$ challenging images with $1,251,642$ annotated people, which is divided into training ($1,201$ images) and testing sets ($334$ images). We compare the proposed method with $6$ state-of-the-art methods (\ie, \cite{DBLP:conf/cvpr/ZhangZCGM16,DBLP:conf/avss/SindagiP17,DBLP:conf/cvpr/SamSB17,DBLP:conf/eccv/IdreesTAZARS18,DBLP:conf/cvpr/WangGL019,DBLP:conf/cvpr/JiangXZZ0D019}) in Table \ref{tab:comparison-results}. Our method achieves $\text{MAE}$ $107.6$ and $\text{MSE}$ $174.8$, surpassing most state-of-the-art methods, which demonstrates that our method produces more accurate density maps.

\subsection{DroneCrowd Dataset}
Besides the above two public datasets, we also evaluate the proposed method on our DroneCrowd dataset for crowd counting, localization and tracking. We report the density map estimation results and speed, \ie, frame-per-second (FPS), of the proposed STANet method and $10$ state-of-the-art methods in Table \ref{tab:density-map-estimation-results}. All codes of the evaluated methods are publicly available or provided by the authors of the corresponding publications\footnote{Since there are no public codes for other video based methods \cite{DBLP:conf/iccv/XiongSY17,DBLP:conf/iccv/ZhangWCM17}, we do not evaluate them on the DroneCrowd dataset.}. All methods are trained on the {\tt training} set and evaluated on the {\tt testing} set. Every $5$ frames are sampled from the video clips in the {\tt training} set to train the evaluated methods.

Meanwhile, we choose the top two previous methods in density map estimation, and post-process the predicted localization maps by finding the local peaks or maximums based on a preset threshold to solve the crowd localization task in Table \ref{tab:crowd-localization-results}. For our STANet method, we directly post-process the predicted localization maps to localize the targets. After that, we use the min-cost flow algorithm \cite{DBLP:conf/cvpr/PirsiavashRF11} to recover the people's trajectories. The evaluation results of the crowd tracking task are shown in Table \ref{tab:crowd-tracking-results}. Some qualitative results are shown in Figure \ref{fig:crowd-visual-results} and more results can be found in the supplementary materials. 

\begin{table*}
\caption{Estimation errors of the density map on the DroneCrowd dataset.}
\vspace{-2mm}
\centering
\setlength{\tabcolsep}{2.5pt}
\scriptsize{
\begin{tabular}{c|c||cc||cc|cc||cc|cc|cc||cc|cc}
\hline
\multirow{2}{*}{Method} &Speed &\multicolumn{2}{c||}{Overall} &\multicolumn{2}{c|}{High} &\multicolumn{2}{c||}{Low} &\multicolumn{2}{c|}{Cloudy} &\multicolumn{2}{c|}{Sunny} &\multicolumn{2}{c||}{Night} &\multicolumn{2}{c|}{Crowded} &\multicolumn{2}{c}{Sparse}\\
\cline{3-18}
&FPS &$\text{MAE}$ &$\text{MSE}$ &$\text{MAE}$ &$\text{MSE}$ &$\text{MAE}$ &$\text{MSE}$ &$\text{MAE}$ &$\text{MSE}$ &$\text{MAE}$ &$\text{MSE}$ &$\text{MAE}$ &$\text{MSE}$ &$\text{MAE}$ &$\text{MSE}$ &$\text{MAE}$ &$\text{MSE}$\\
\hline
MCNN \cite{DBLP:conf/cvpr/ZhangZCGM16} &${\bf 28.98}$ &$34.7$ &$42.5$ &$36.8$ &$44.1$ &$31.7$ &$40.1$ &$21.0$ &$27.5$ &$39.0$ &$43.9$ &$67.2$ &$68.7$ &$29.5$ &$35.3$ &$37.7$ &$46.2$\\
C-MTL \cite{DBLP:conf/avss/SindagiP17}&$2.31$ &$56.7$ &$65.9$ &$53.5$ &$63.2$ &$61.5$ &$69.7$ &$59.5$ &$66.9$ &$56.6$ &$67.8$ &$48.2$ &$58.3$ &$81.6$ &$88.7$ &$42.2$ &$47.9$\\
MSCNN \cite{DBLP:conf/icip/ZengXCQZ17}&$1.76$ &$58.0$ &$75.2$ &$58.4$ &$77.9$ &$57.5$ &$71.1$ &$64.5$ &$85.8$ &$53.8$ &$65.5$ &$46.8$ &$57.3$ &$91.4$ &$106.4$ &$38.7$ &$48.8$\\
LCFCN \cite{DBLP:conf/eccv/LaradjiRPVS18}&$3.08$ &$136.9$ &$150.6$ &$126.3$ &$140.3$ &$152.8$ &$164.8$ &$147.1$ &$160.3$ &$137.1$ &$151.7$ &$105.6$ &$113.8$ &$208.5$ &$211.1$ &$95.4$ &$110.0$\\
SwitchCNN \cite{DBLP:conf/cvpr/SamSB17}&$0.014$ &$66.5$ &$77.8$ &$61.5$ &$74.2$ &$74.0$ &$83.0$ &$56.0$ &$63.4$ &$69.0$ &$80.9$ &$92.8$ &$105.8$ &$67.7$ &$79.8$ &$65.7$ &$76.7$\\
ACSCP \cite{DBLP:conf/cvpr/ShenXNWHY18}&$1.58$ &$48.1$ &$60.2$ &$57.0$ &$70.6$ &$34.8$ &$39.7$ &$42.5$ &$46.4$ &$37.3$ &$44.3$ &$86.6$ &$106.6$ &$36.0$ &$41.9$ &$55.1$ &$68.5$\\
AMDCN \cite{DBLP:conf/cvpr/DebV18}&$0.16$ &$165.6$ &$167.7$ &$166.7$ &$168.9$ &$163.8$ &$165.9$ &$160.5$ &$162.3$ &$174.8$ &$177.1$ &$162.3$ &$164.3$ &$165.5$ &$167.7$ &$165.6$ &$167.8$\\
CSRNet \cite{DBLP:conf/cvpr/LiZC18}&$3.92$ &$19.8$ &$25.6$ &$17.8$ &$25.4$ &$22.9$ &$25.8$ &$12.8$ &$16.6$ &$19.1$ &$22.5$ &$42.3$ &$45.8$ &$20.2$ &$24.0$ &$19.6$ &$26.5$\\
StackPooling \cite{DBLP:journals/corr/abs-1808-07456}&$0.73$ &$68.8$ &$77.2$ &$68.7$ &$77.1$ &$68.8$ &$77.3$ &$66.5$ &$75.9$ &$74.0$ &$83.4$ &$65.2$ &$67.4$ &$95.7$ &$101.1$ &$53.1$ &$59.1$\\
DA-Net \cite{DBLP:journals/access/ZouSQZ18}&$2.52$ &$36.5$ &$47.3$ &$41.5$ &$54.7$ &$28.9$ &$33.1$ &$45.4$ &$58.6$ &$26.5$ &$31.3$ &$29.5$ &$34.0$ &$56.5$ &$68.3$ &$24.9$ &$28.7$\\
\hline
\hline
STANet (w/o ms)&$9.49$ &$26.3$ &$31.4$ &$27.3$ &$33.9$ &$24.7$ &$27.1$  &$21.3$ &$23.2$ &$29.5$ &$37.7$  &$34.7$ &$38.0$ &$22.4$ &$25.0$  &$28.5$ &$34.5$\\
STANet (w/o loc)&$3.03$ &$17.9$ &$20.7$ &$17.8$ &$20.9$ &$17.9$ &$20.5$ &$16.3$ &$18.9$ &$20.5$ &$23.8$  &$17.3$ &$19.3$ &${\bf 17.3}$ &${\bf 20.2}$ &$18.2$ &$21.0$\\
STANet (w/o ass)&$2.83$ &$16.8$ &$21.8$ &${\bf 16.3}$ &$20.6$ &${\bf 15.9}$ &$23.6$ &${\bf 10.5}$ &${\bf 14.2}$ &$22.4$ &$27.3$  &$20.3$ &$27.6$ &$21.3$ &$26.8$ &${\bf 13.0}$ &${\bf 18.3}$\\
STANet &$2.74$ &${\bf 16.7}$ &${\bf 19.8}$ &$17.0$ &${\bf 20.0}$ &$16.4$ &${\bf 19.5}$ &$14.3$ &$17.2$ &${\bf 19.0}$ &${\bf 21.1}$ &${\bf 19.6}$ &${\bf 24.1}$  &$19.4$ &$22.1$ &$15.2$ &$18.4$\\
\hline
\end{tabular}}
\label{tab:density-map-estimation-results}
\end{table*}

\begin{figure*}[t]
\centering
\includegraphics[width=\linewidth]{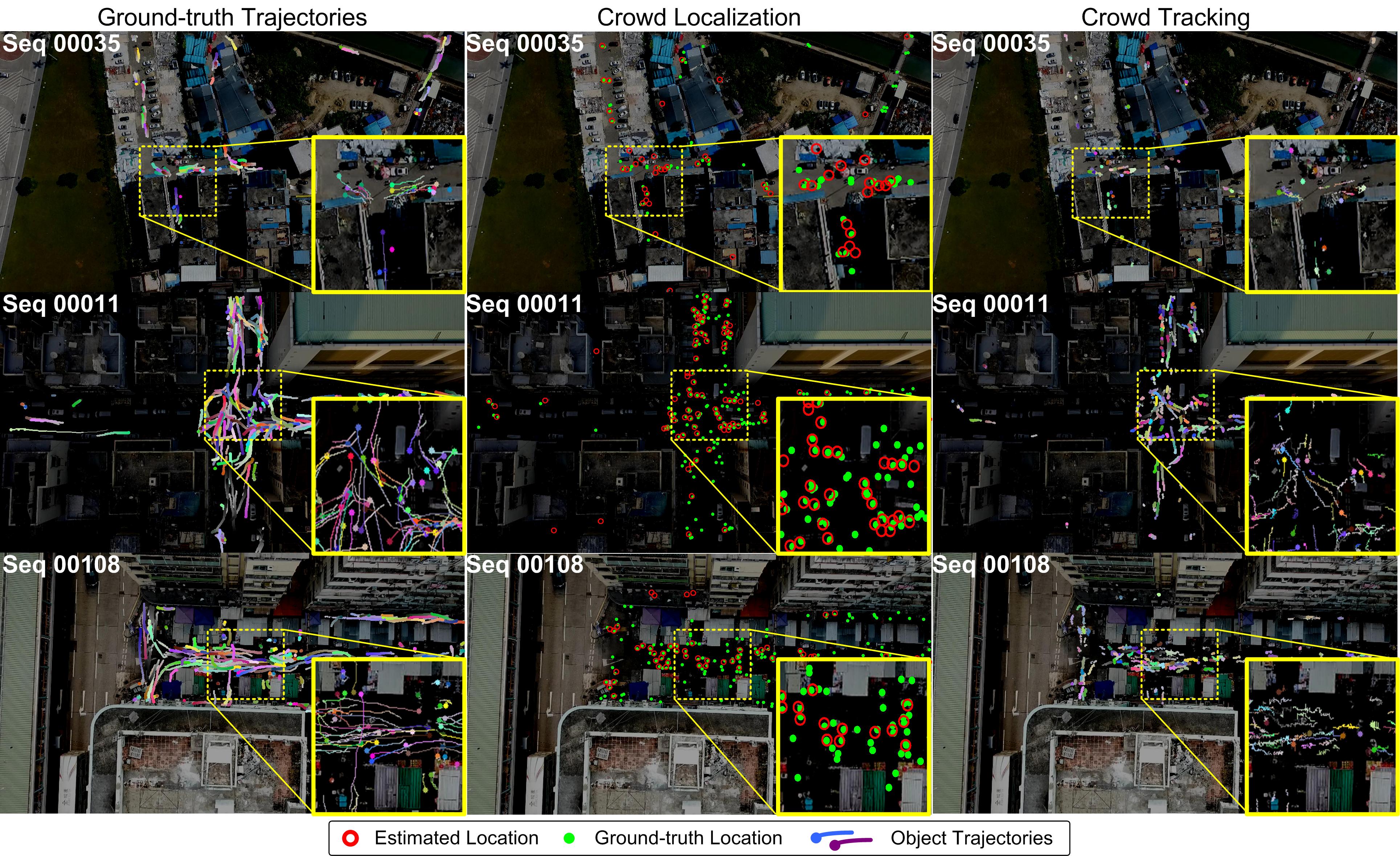}
\caption{Qualitative results of STANet on three sequences in our DroneCrowd. Best view in color version.}
\label{fig:crowd-visual-results}
\end{figure*}

{\flushleft {\bf Density map estimation}.}
As shown in Table \ref{tab:density-map-estimation-results}, our STANet performs favorably against the state-of-the-art methods, with an improvement of $3.1$ $\text{MAE}$ and $5.8$ $\text{MSE}$ in comparison to the second best CSRNet \cite{DBLP:conf/cvpr/LiZC18} in the overall {\tt testing} set, \ie, $16.7$ $\text{MAE}$ of STANet {\em vs.} $19.8$ $\text{MAE}$ of CSRNet \cite{DBLP:conf/cvpr/LiZC18}, and $19.8$ $\text{MSE}$ of STANet {\em vs.} $25.6$ $\text{MSE}$ of CSRNet \cite{DBLP:conf/cvpr/LiZC18}. The third best MCNN \cite{DBLP:conf/cvpr/ZhangZCGM16} achieves the third best $\text{MAE}$ score of $34.7$ with the speed of $29.98$ FPS. This suggests that our method generates more accurate and robust density maps in different scenarios. 

To further analyze the results, we report the performance on several break-down subsets based on the video-level attributes, \ie, the {\em High} and {\em Low} subsets based on the {\em Altitude} attribute, the {\em Cloudy}, {\em Sunny}, and {\em Night} subsets based on the {\em Illumination}, and the {\em Crowd} and {\em Sparse} subsets based on the {\em Density} attribute. As shown in the sixth column of Table \ref{tab:density-map-estimation-results}, LCFCN \cite{DBLP:conf/eccv/LaradjiRPVS18} and AMDCN \cite{DBLP:conf/cvpr/DebV18} fail to perform well in the {\em Crowd} subset, producing the two worst $\text{MAE}$ and $\text{MSE}$ scores, \ie, $208.5$ and $165.5$ $\text{MAE}$, and $211.1$ and $167.7$ $\text{MSE}$, respectively. We speculate that this may be due to several reasons. LCFCN \cite{DBLP:conf/eccv/LaradjiRPVS18} uses a loss function to encourage the network to output a segmentation blob for each object in crowd counting. However, in crowd scenarios, each object may contain only few pixels, making it difficult to separate objects accurately. AMDCN \cite{DBLP:conf/cvpr/DebV18} uses multiple columns of large dilation convolution operations, which inevitably integrate considerable background noise, affecting the accuracy in density map estimation. In contrast, MCNN \cite{DBLP:conf/cvpr/ZhangZCGM16} uses multi-column CNNs to learn the features adaptive to variations in object size due to perspective effect or image resolution. Meanwhile, STANet achieves the best result by gradually applying the attention module on the combined multi-scale feature maps. This phenomenon strongly demonstrates the effectiveness and importance of exploiting multi-scale features in density map estimation.

\begin{table}[t]
\centering
\caption{Crowd localization accuracy on DroneCrowd.}
\vspace{-2mm}
\setlength{\tabcolsep}{2.0pt}
\footnotesize{
\begin{tabular}{c|c|ccc}
\hline
Methods &L-mAP &L-AP$@10$ &L-AP$@15$ &L-AP$@20$ \\
\hline
MCNN-L &$9.05\%$ &$9.81\%$ &$11.81\%$ &$12.83\%$ \\
CSRNet-L &$14.40\%$ &$15.13\%$ &$19.77\%$ &$21.16\%$ \\
\hline\hline
STANet-L (w/o ms)&$15.19\%$  &$15.64\%$ &$21.12\%$ &$22.64\%$ \\
STANet-L (w/o loc)&$27.36\%$  &$29.17\%$ &$33.70\%$ &$36.27\%$ \\
STANet-L (w/o ass)&$27.96\%$  &$30.33\%$ &$35.80\%$ &$38.04\%$ \\
STANet-L &${\bf 28.43\%}$  &${\bf 30.53\%}$ &${\bf 36.33\%}$ &${\bf 39.12\%}$ \\
\hline
\end{tabular}}
\label{tab:crowd-localization-results}
\end{table}

To study the influence of each module in STANet, we construct three variants and evaluate them on the DroneCrowd dataset, \ie, STANet (w/o ass), STANet (w/o loc), and STANet (w/o ms), in Table \ref{tab:density-map-estimation-results}. Specifically, for a fair comparison, we use the same parameter settings and input size in evaluation. All variants are trained on the {\tt training} set and tested on the {\tt testing} set. STANet (w/o ass) indicates the method that removes the association head from STANet. STANet (w/o loc) indicates the method that removes the localization head from STANet (w/o ass). STANet (w/o ms) denotes the method that further removes the multi-scale features in prediction, \ie, only using the first four groups of convolution layers in VGG16.

As shown in Table \ref{tab:density-map-estimation-results}, our STANet achieves better results than its variants. After removing the association head, the $\text{MSE}$ score on overall set increases $2.0$ ($19.8$ {\em vs.} $21.8$), demonstrating that temporal association helps improve the robustness. If we remove the localization head, the errors increase $1.1$ in $\text{MAE}$ ($16.8$ {\em vs.} $17.9$). This performance drop validates the importance of the localization head. If we further remove the multi-scale feature module, the $\text{MAE}$ score is increased from $17.9$ to $26.3$. This sharp decline (\ie, $8.4$) in accuracy demonstrates that the multi-scale features significantly promote the performance in density map estimation. In addition, we notice that STANet (w/o ass) performs better than STANet in the {\em Cloudy} subset. We speculate that this phenomenon is caused by the inaccuracy of temporal information in cloudy scenarios. The similar cases are also observed in the {\em Sparse} subset.

{\flushleft {\bf Crowd localization}.}
We conduct non-maximal suppression to localize people's heads in videos. Specifically, we find the local peaks or maximums based on a preset threshold on the predicted localization map in each frame. The crowd localization results of the two methods with top results (\ie, MCNN \cite{DBLP:conf/cvpr/ZhangZCGM16}, CSRNet \cite{DBLP:conf/cvpr/LiZC18}) in density map estimation and our STANet variants are shown in Table \ref{tab:crowd-localization-results}, named as MCNN-L, CSRNet-L, STANet-L (w/o ms), STANet-L (w/o loc), STANet-L (w/o ass), and STANet-L, respectively. 

As shown in Table \ref{tab:crowd-localization-results}, STANet achieves the best localization accuracy of $28.43\%$ L-mAP, surpassing the second best CSRNet \cite{DBLP:conf/cvpr/LiZC18} $14.03\%$ L-mAP. It indicates that our method is not only able to predict the distributions of objects in the scenes, but also generates relatively more accurate localizations of each object instance. Without the association head, the L-mAP score decreases $0.47\%$, indicating that temporal coherence helps improve the localization accuracy slightly. If we remove both association and localization heads, the L-mAP score decreases $0.60\%$. It demonstrates that the localization head enforces the network to focus on more discriminative features to localize the people's heads. If we further remove the multi-scale feature design, the L-mAP score drops from $27.36\%$ to $15.19\%$, which validates that multi-scale features play a critical role on the performance in crowd localization.

{\flushleft {\bf Crowd tracking}.}
Moreover, we also evaluate the crowd tracking results on our DroneCrowd dataset in Table \ref{tab:crowd-tracking-results}. Specifically, we construct six crowd tracking methods, \ie MCNN-T, CSRNet-T, STANet-T (w/o ms), STANet-T (w/o loc), STANet-T (w/o ass), and STANet-T, using the min-cost flow method \cite{DBLP:conf/cvpr/PirsiavashRF11} to associate the location points generated by the corresponding crowd localization methods.

As shown in Table \ref{tab:crowd-tracking-results}, we notice that our STANet-T produces the best results with the top T-mAP score, \ie, $23.76\%$, that is $14.08\%$ higher than the second best method CSRNet-T. STANet-T (w/o ass) produces comparable results with STANet-T ($23.76\%$ vs. $22.76\%$). We speculate that the association head is effective to use temporal association information to recover the trajectories of people. The T-mAP score of STANet-T (w/o loc) decreases $1.63\%$ compared with STANet-T (w/o ass), and STANet-T (w/o ms) only achieves T-mAP score $10.58\%$. These results indicate that association and localization heads and multi-scale representation are critical in crowd tracking. However, all the results are still far from satisfactory in real applications. It indicates that DroneCrowd is extremely challenging for crowd tracking and much effort is needed to develop more effective methods in real scenarios.

\begin{table}[t]
\centering
\caption{Crowd tracking accuracy on DroneCrowd.}
\vspace{-2mm}
\setlength{\tabcolsep}{2.0pt}
\footnotesize{
\begin{tabular}{c|c|ccc}
\hline
Methods   &T-mAP &T-AP$@0.10$ &T-AP$@0.15$ &T-AP$@0.20$ \\
\hline
MCNN-T  &$7.49\%$  &$11.36\%$ &$6.73\%$ &$4.39\%$ \\
CSRNet-T   &$9.68\%$  &$17.55\%$ &$7.80\%$ &$3.71\%$ \\
\hline\hline
STANet-T (w/o ms)   &$10.58\%$  &$19.06\%$ &$8.58\%$ &$4.09\%$ \\
STANet-T (w/o loc) &$21.15\%$  &$30.54\%$ &$21.35\%$ &$14.57\%$ \\
STANet-T (w/o ass) &$22.78\%$  &${\bf 31.39\%}$ &$22.07\%$ &$14.90\%$ \\
STANet-T  &${\bf 23.76\%}$  &$30.96\%$ &${\bf 23.05\%}$ &${\bf 17.28\%}$ \\
\hline
\end{tabular}}
\label{tab:crowd-tracking-results}
\end{table}

\section{Conclusion}
In this work, we propose the STANet method to jointly solve density map estimation, localization, and tracking in crowds of video clips captured by drones. To better evaluate the performances of density map estimation, localization, and tracking on drones, we collect and annotate a new dataset, DroneCrowd, which consists of $112$ video clips with $33,600$ high resolution frames and more than $4.8$ million head annotations. This is the largest dataset to date in terms of annotated trajectories of heads for density map estimation, crowd localization, and tracking on drones. Our model performs favorable against the state-of-the-art crowd counting methods on the three challenging datasets, demonstrating its effectiveness. We hope the dataset and the proposed method can facilitate the research and development in crowd counting, localization and tracking on drones.

{\small
\bibliographystyle{ieee_fullname}
\bibliography{reference}
}

\end{document}